\documentclass{ifacconf}

\usepackage{graphicx}      
\usepackage{natbib}        
\usepackage{amsmath}
\usepackage{amsfonts}
\usepackage{amssymb}
\newcommand{\norm}[2]{\left\Vert #1 \right \Vert_{#2}}

\begin{document}
\begin{frontmatter}

\title{Optimal path planning of multi-agent cooperative systems with rigid formation} 

\author[First]{Ananda Rangan Narayanan,} 
\author[First]{Mi Zhou,}
\author[First]{Erik Verriest} 
\address[First]{Georgia Institute of Technology, 
   Atlanta, GA 30332 USA (e-mail:  anarayanan70@gatech.edu, mzhou91@gatech.edu, erik.verriest@ece.gatech.edu).}

\begin{abstract}                
In this article, we consider the path-planning problem of a cooperative homogeneous robotic system with rigid formation.
An optimal controller is designed for each agent in such rigid systems based on Pontryagin's minimum principle theory.
We found that the optimal control for each agent is equivalent to the optimal control for the Center of Mass (CoM).
This equivalence is then proved by using some analytical mechanics. 
Three examples are finally simulated to illustrate our theoretical results.
One application could be utilizing this equivalence to simplify the original multi-agent optimal control problem.
\end{abstract}

\begin{keyword}
scalable optimal control, rigid robotic system, multi-agent system, homogeneous, cooperative.
\end{keyword}

\end{frontmatter}

\section{Introduction}
The control of a rigid system lies in human being's everyday life.
Examples include self-balancing scooter systems like Hoverboards, car backing into a garage, multi-robot system steering with rigid formation, furniture moving\cite{VERRIEST}, multi-agent rigid formation keeping~\cite{rigidformation}, and so on.
This article concerns the optimal path planning of multi-agent systems with rigid formation.
In \cite{VERRIEST}, the authors solved the problem of moving a line segment with minimal lengths of the two paths described by the endpoints.
Extended from \cite{VERRIEST}, which considered moving a rod in such a way that the sum of distances traveled by both endpoints is minimized, we consider the energy cost for the two endpoints.
Our problem can be regarded as a multi-agent system where between every two agents, there is a rigid rod.
Different from \cite{VERRIEST}, we optimize a control energy function and prove that the optimal control for each agent is equivalent to the optimal control for the center of mass.
It belongs to the area of scalable optimal control.
Instead of seeing each robot as an individual, we utilize the analytical mechanic's method that will simplify our problem significantly.

The analytical approach to the multi-agent motion problem views the agent not as an isolated unit but as a part of a mechanical system which is an assembly of agents that interact with each other ~\cite{goldstein}.
Using this approach can significantly reduce the complexity of analyzing the motion of individual particles.
We introduce this concept to solving the optimal control for homogeneous multi-agent system problems, which will greatly reduce the dimension of the state space and simplify this multi-agent optimal control problem.

We believe the results of this research work can be applied in path planning for robotic vehicles in some scenarios.
Moreover, this work can also provide some insight into controlling multi-agent systems, such as drone flight shows, and parallel robots, such as Stewart platform.
Parallel robots can be defined as closed-loop mechanisms composed of end-effectors having $n$ degrees of freedom and a fixed base \cite{Merlet2008}.
In this mechanism, the control of the end-effectors is constrained by the rigidity of the base, which is similar to our problem.

This article is constructed as follows: in Section \ref{sec:problemformulation}, we first present some preliminaries related to analytical mechanics and then formulate our multi-agent optimal control problem; Section \ref{sec:theoryresults} is our derivation of the optimal controller; then in Section \ref{sec: simulations}, simulation results for three multi-agent systems are given to illustrate our theoretical results; Section \ref{sec:conclusion} concludes this article and some future works.

\section{Preliminaries and Problem formulation} \label{sec:problemformulation}
\subsection{Preliminaries in analytical mechanics}
Analytical mechanics was developed during the 18th century after Newtonian mechanics.
Different from Newtonian mechanics, analytical mechanics take advantage of a system's constraints which limit the degrees of freedom the system can have, and use scalar properties of motion representing the system as a whole, such as its total kinetic energy and potential energy.
This method can reduce the number of coordination needed to solve the motion.
In the following, we will present some definitions and facts in analytical mechanics.
\begin{defn} [\cite{goldstein}]
Usually, a rigid body can be defined as a system of particles in which the distance $r_{ij}$ are fixed and can not vary with time.  
\end{defn}
\begin{defn} [\cite{goldstein}]
If the conditions of constraint can be expressed as equations connecting the coordinates of the particles having the form $f(r_1, r_2, r_3,...,t)=0$, then the constraints are said to be \textbf{holonomic}.
\end{defn}
\begin{fact}
The rigid body has holonomic constraints.
\end{fact}
\begin{thm}[Chasles' theorem]
Any general displacement of a rigid body can be represented by a translation plus a rotation.
\end{thm}
This theorem suggests that we can split the problem of rigid body motion into two separate phases, one concerned solely with the translational motion of the body, and the other, with its rotational motion.
This lays the foundation for our proposed method.
\subsection{Problem formulation}
\begin{figure}[!htp]
    \centering
    \includegraphics[width=\linewidth]{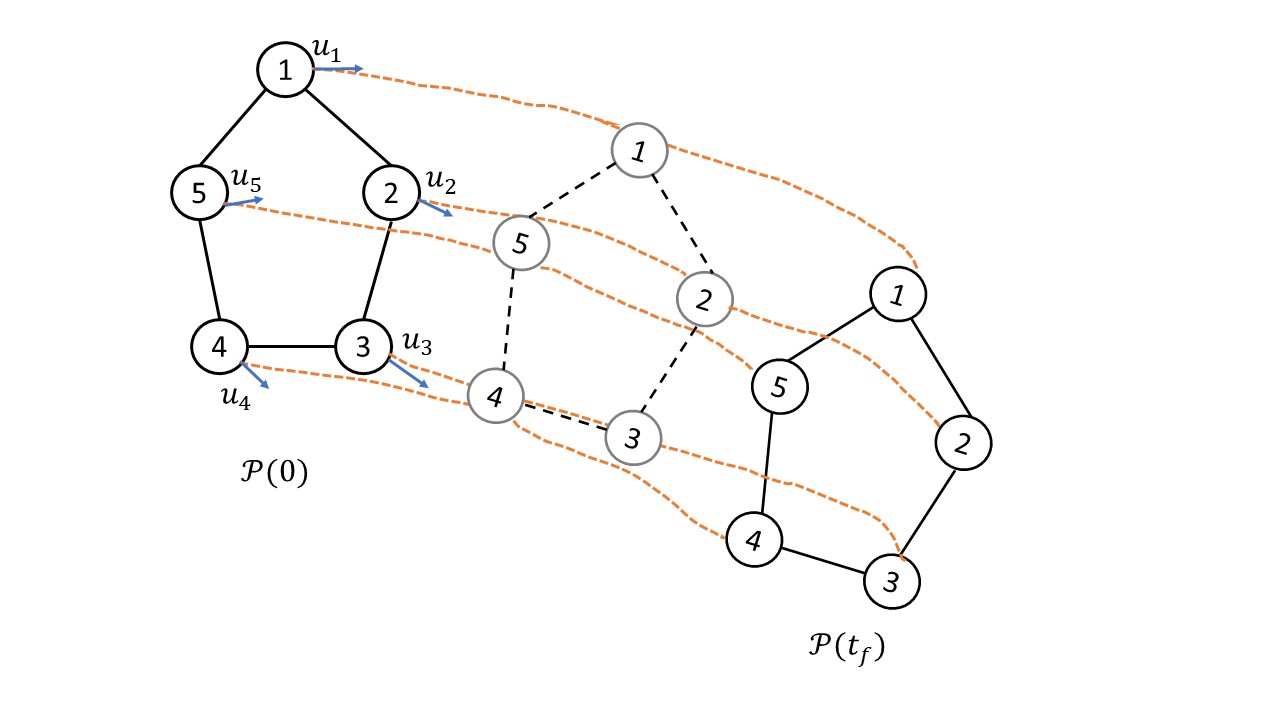}
    \caption{Problem description: a five-agent example}
    \label{fig:problemdesc}
\end{figure}

Consider a system composed of $N$ agents.
Each agent has dynamics $\dot r_i = f(r_i,u_i), \; i=1,2,...,N$.
The objective is to regard the whole robotic system as a rigid system and steer it optimally from position $\mathcal{P}(0) = \{r_1(0), r_2(0), \dots, r_N(0)\}$ to $\mathcal{P}(t_f)= \{r_1(t_f), r_2(t_f), \dots, r_N(t_f)\}$ while keeping their formation $\mathcal{S}(t)=\{r_{ij}(t) = r_{ij}(0) | \forall i,j \in 1,\dots, N \}$ the same during the whole time period.
One example is multi-agent systems cooperatively moving a big object from one position to another position while each agent has its all intelligence.
As shown in Fig. \ref{fig:problemdesc}, five agents are at initial position $\mathcal{P}(0)$ and have initial formation $\mathcal{S}(0)$ at $t=0$. 
Each agent $i$ has its own controller $u_i$ and they want to move cooperatively to position $\mathcal{P}(t_f)$ while keeping the relative position of each agent in this system the same.
With these definitions and objectives, we can thus formulate a Bolza-type optimal control problem with some equality constraints:
\begin{align}
    \min J =  \phi(r(t_f))+\int_0^{t_f}\sum_i L_i(r_i,u_i) dt \label{eqn:performanceInd}\\
    s.t., \;
    \dot r_i = f_i(r_i,u_i), \; i=1,2,...,N \label{eqn:dynamics}\\
    \mathcal{S}(0)=\mathcal{S}(t_f) = \mathcal{S}(t),\; \forall t \in [0, t_f] \label{eqn:constraints}
\end{align}
where $r_i  \in \mathbb{R}^2$ is the state, $u_i \in \mathcal{U}$ is the control input,  $t_f$ denotes the terminal time.
The system dynamics $f_i:\mathbb{R}^n\times \mathcal{U}\rightarrow \mathbb{R}^n$ is a continuous and differentiable function.
The stage cost $L_i(r_i,u_i):\mathbb{R}^n\times \mathcal{U}\rightarrow \mathbb{R}$ is also continuous and differentiable.
With these assumptions, we can use Pontryagin minimum principle to obtain the optimality conditions and the Euler-Lagrangian equation, thus solving this optimal control problem for each agent.
This is well-developed already so our aim here is to introduce a new analytical-mechanics-based method to simplify this multi-state-multi-controller problem.

\section{Theoretic results}\label{sec:theoryresults}
In this section, we will introduce our method to make the original multi-agent system optimal control problem equivalent to a single-agent optimal control problem.
We first derive the optimality conditions for the two-agent case with a simple position-velocity model.
Multi-agent cases are then extended using the same logic.
\subsection{Two-agent system}
To make it more clear, we first provide a two-agent system in a two-dimensional plane as an example.
Consider the following two-agent system:
\begin{align}\label{eqn:twoDyn}
\dot r_i = u_i, \; i=1,2
\end{align}
where $r_i \triangleq [x,y]^\top$, $u_i \in \mathbb{R}^2$.
The performance index is defined as the kinetic energy expended:
\begin{align}\label{eqn:twoCost}
J = \frac{1}{2}\int_0^{t_f} \norm{u_1}{2}^2+\norm{u_2}{2}^2 \rm{d}t.
\end{align}
To make the two agents keep the same distance during the whole process, the following rigidity constraint should be added:
\begin{align} \label{eqn:rididCons}
\norm{r_1-r_2}{2} = l 
\end{align}
where $l$ is a constant denoting the distance between two agents.

Construct the following Hamiltonian
\begin{align}
H = \frac{1}{2}\norm{u_1}{2}^2+\frac{1}{2}\norm{u_2}{2}^2+\lambda_1^\top u_1 +\lambda_2^\top u_2 +\frac{1}{2}\mu \norm{r_1-r_2}{2}^2
\end{align}
where $\lambda_i \in \mathbb{R}^2$ is the Lagrangian multiplier, $\mu$ is also a multiplier corresponding to the equality constraints \eqref{eqn:rididCons}. 
Applying the Pontryagin Minimum Principle, we get equations for the optimal control $u_i$ and the dynamics of the Lagrange multipliers $\lambda_i$.  
The optimality conditions yield:
\begin{align*}
\frac{\partial H}{\partial u_i} = 0 \implies u_i + \lambda_i = 0, \;  i = 1,2.
\end{align*}
The co-state equations yield:
\begin{align*}
\dot{\lambda}_i = -\frac{\partial H}{\partial r_i} \implies \dot{\lambda}_i = -\mu(r_i - r_j), \; (i,j) = \{(1,2),(2,1)\}.
\end{align*}
Hence, 
\begin{align*}
\ddot{r}_i = \mu(r_i - r_j), \; (i,j) = \{(1,2),(2,1)\}. 
\end{align*}
This leads to
\begin{align}
\ddot{r}_1 + \ddot{r}_2 &= 0,  \label{eqn:secDeriSum}\\
\ddot{r}_1 - \ddot{r}_2 &= 2\mu(r_1 - r_2) .\label{eqn:secDeriDed}
\end{align}
Define
\begin{align} \label{eqn:COMdefine}
r_c = \frac{r_1+r_2}{2}.    
\end{align}
Eqn. \eqref{eqn:secDeriSum} thus implies that the $\ddot{r}_c = 0$.
In other words, the center of mass does not accelerate.\\
Rewriting the coordinates of the agents $r_i$ in terms of the center of mass,
\begin{align}
r_1 = r_c + \frac{l}{2}\hat{s}(\theta) \label{eqn:r1com}, \\
r_2 = r_c - \frac{l}{2}\hat{s}(\theta) \label{eqn:r2com}.
\end{align}
where $r_c$ is the coordinate of the center of mass as defined in Eqn.\eqref{eqn:COMdefine} and $\hat{s}(\theta)$ is a unit vector at an angle $\theta$ from the horizontal.
\begin{align*}
\hat{s}(\theta) = \begin{bmatrix}
\cos(\theta) \\
\sin(\theta)
\end{bmatrix}.    
\end{align*}
This representation of $r_i$ automatically satisfies the constraint equation \eqref{eqn:rididCons}. Taking the translational and angular velocity of the center of mass as $u_c$ and $\omega_c$ respectively, velocities of the agents $r_i$ can be represented as,
\begin{align}\label{eqn:twoAgentControl}
\left\{\begin{matrix}
u_1 &= \dot{r}_1 = u_c + \frac{l\omega}{2}\hat{s}^{\bot}(\theta) \\
u_2 &= \dot{r}_2 = u_c - \frac{l\omega}{2}\hat{s}^{\bot}(\theta)
\end{matrix}\right.  
\end{align}
where $\hat{s} \hat{s}^\perp(\theta)=0$.

The performance index \eqref{eqn:twoCost} can then be rewritten as,
\begin{align}
    J = \frac{1}{2}\int_0^{t_f}2\norm{u_c}{2}^2 + \frac{l^2\omega^2}{2} \rm d t.
\end{align}
Thus the optimal trajectory has the center of mass traveling with constant translation and rotational velocity
\begin{align}\label{eqn:optimalSoln}
    u_c = \frac{r_c(t_f) - r_c(0)}{t_f},\;
    \omega = \frac{\theta(t_f) - \theta(0)}{t_f},
\end{align}
which is obvious from Pontryagin's minimum principle.

\subsection{Multi-agent systems}
The results from the above two-agent system can be extended to a multi-agent system of $N$ agents.
\begin{lem}
    Planar rigidity constraints: For an $N$-agent planar system to be rigid, there must be at least $2N-3$ constraint equations.
\end{lem}
\begin{pf}
A rigid planar system has three degrees of freedom, typically represented by the Cartesian coordinates and the heading angles.
A point $r_i$ in $\mathbb{R}^2$ has two degrees of freedom, generally denoted by its Cartesian coordinates.
Thus, for $N$ points to form a rigid system, there must be at least $2N-3$ constraints, leaving exactly 3 degrees of freedom. 
\end{pf}

\begin{figure}
    \centering
    \includegraphics[width=\linewidth]{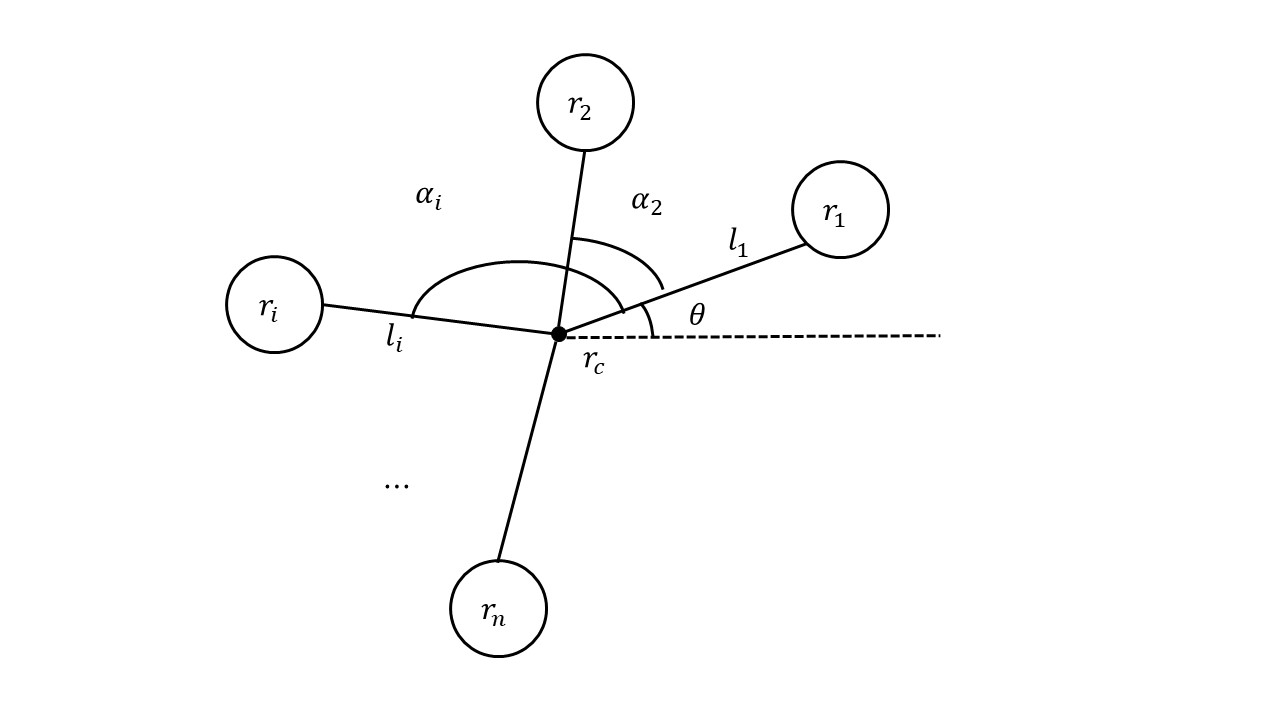}
    \caption{Illustrating the definition of notations}
    \label{fig:notations}
\end{figure}

Let us define the system of $N$ agents by their center of mass and a heading angle. 
\begin{align}
    \mathcal{P}(r_c,\theta) = \begin{cases}
    r_1 = r_c + l_1\hat{s}(\theta) \\
    r_i = r_c + l_i\hat{s}(\alpha_i+\theta), \;  i = 2,\dots,n-1 \\
    r_N = r_c + \sum_{i=1}^{n-1} (r_c - r_i)
    \end{cases}.
\end{align}
As shown in Fig. \ref{fig:notations},
the first point is at a fixed distance $l_1$ from $r_c$ and chosen to match the heading $\theta$. The points $r_i$ for $i=2,\dots,N-1$ are at a distance $l_i$ from $r_c$ and the angle between $r_i$ and $r_1$ is fixed with respect to the center of mass, $\angle r_i r_c r_1 = \alpha_i$. 
The last point $r_N$ is chosen in such a way that the center of mass of the resulting system lies at $r_c$. 
This representation automatically satisfies the $2N-3$ constraints for a rigid system of $N$ agents since
\begin{align*}
    \norm{r_i - r_c}{2} = l_i , \; i= 1,2,\dots,N-1
\end{align*}
and 
\begin{align*}
\langle r_i - r_c, r_1 - r_c \rangle = d_1 d_i \cos(\alpha_i), \; i = 2,\dots,N-1.
\end{align*}
Taking $\alpha_1 = 0$ the velocity of the agents $r_i$ can now be represented as,
\begin{align*}
    u_i = \dot{r}_i &= u_c + l_i\omega\hat{s}^{\bot}(\alpha_i + \theta), \; i = 1,\dots, N-1\\
    u_N = \dot{r}_N &= u_c - \sum_{i=1}^{N-1}l_i\omega\hat{s}^{\bot}(\alpha_i+\theta)
\end{align*}
The original performance metric can then be rewritten as 
\begin{align*}
    J = \frac{1}{2}\int_0^{t_f}N\norm{u_c}{2}^2 &+ \\ \omega^2( \sum_{i=1}^{N-1}{l_i^2} &+ \sum_{i=1}^{N-1}\sum_{j=1}^{N-1}l_il_j \cos(\alpha_i-\alpha_j)) \rm d t.
\end{align*}

In other words, the total energy of the $N$-agents rigid system can be rewritten as the sum of translational energy and rotational energy of the center of mass. 
Thus we transform the original multi-agent optimal control problem to a single-agent (i.e., the center of mass) optimal control problem with another performance index function. 
Hence the solution from \eqref{eqn:optimalSoln} holds for the multi-agent case too.
\section{Numerical Applications} \label{sec: simulations}
In this section, we provide three examples to illustrate our theoretical results.
The first example is a two-agent system called the pipe model.
Then the second example is a three-agent system.
The last example is a four-agent linear system.
\subsection{Example 1: Pipe model with linear dynamics}
\begin{figure}[!htp]
    \centering
    \includegraphics[width=\linewidth]{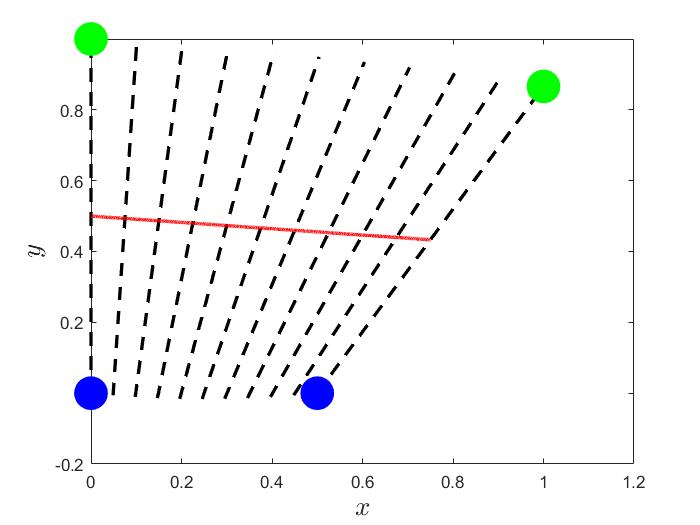}
    \caption{Trajectory: filled dot: two people; black dotted line: the trajectory of this two-people-pipe system; red line: the trajectory of the center of mass}
    \label{fig:twoAgentTraj}
\end{figure}

\begin{figure}[!htp]
    \centering
    \includegraphics[width=\linewidth]{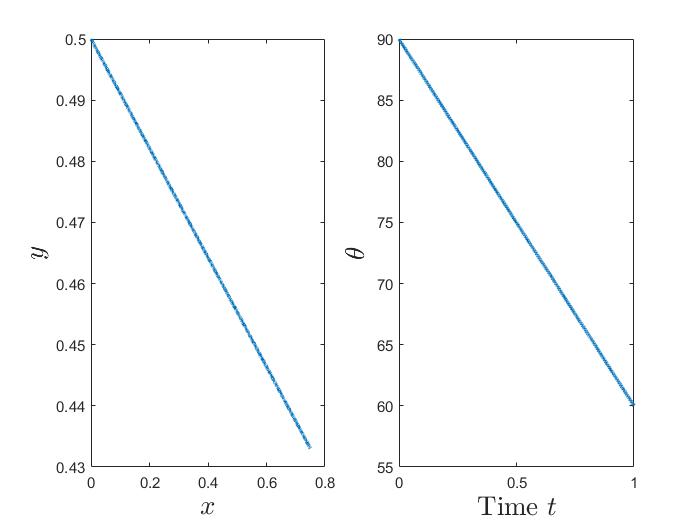}
    \caption{Trajectory of CoM; Angle of CoM (unit: degrees)}
    \label{fig:two_agent_CM}
\end{figure}

Imagine two people moving a pipe cooperatively from one position to another in a fixed time interval. 
The question is how each person's movement can make the energy cost minimal.
Eqn. \eqref{eqn:twoDyn}, \eqref{eqn:twoCost}, \eqref{eqn:rididCons} had combined this problem in mathematics language.
Here we will find the optimal solution with a specific example.
The initial position is set as $[0,0]^\top$, $[0,1]^\top$ for two agents respectively.
The terminal position of the two agents is respectively $[\frac{1}{2},0]^\top$ and $[1, \frac{\sqrt{3}}{2}]^\top$.
The initial time is $t_0=0$ and the terminal time is $t_f = 1$.
Solving this problem optimally, we obtain the optimal trajectory for each agent and the CoM as shown in Fig. \ref{fig:twoAgentTraj}.
The trajectory of this two-mountaineers-pipe model includes a rotation and a sliding at the same time.
The corresponding optimal cost is $J= 0.6355$.
In this problem, the optimal control of the center of mass is indeed a constant value, which makes the center of mass steer from the initial position to the terminal position in a line, as shown in the red line in Fig. \ref{fig:twoAgentTraj}.
Fig. \ref{fig:two_agent_CM} shows the trajectory of the CoM and the angle $\theta$ of the mountaineer-pipe system with respect to the global coordinates, which shows that the CoM moves in a constant velocity and constant angular velocity $\omega$ as suggested in Eqn.~\eqref{eqn:optimalSoln}.
\subsection{Example 2: three-agents with irregular shape}
\begin{figure}[!htp]
    \centering
    \includegraphics[width=\linewidth]{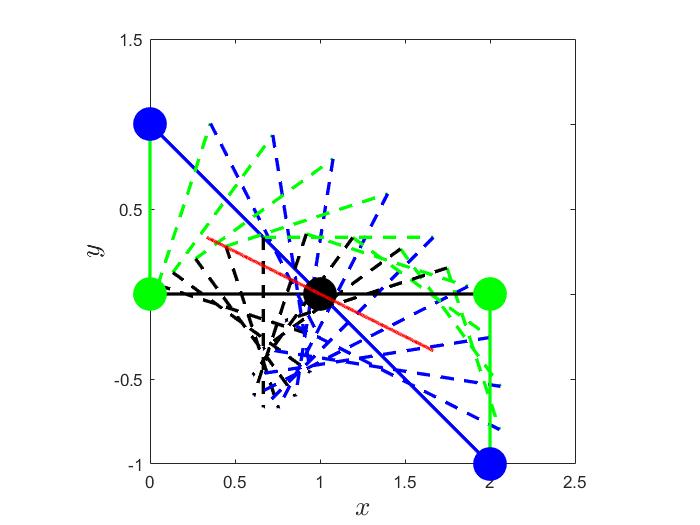}
    \caption{Trajectory: filled dot: three agents, each color represents an agent; corresponding dotted line: the trajectory of each agent; red line: the trajectory of the center of mass}
    \label{fig:threeagent_traj}
\end{figure}

\begin{figure}[!htp]
    \centering
    \includegraphics[width=\linewidth]{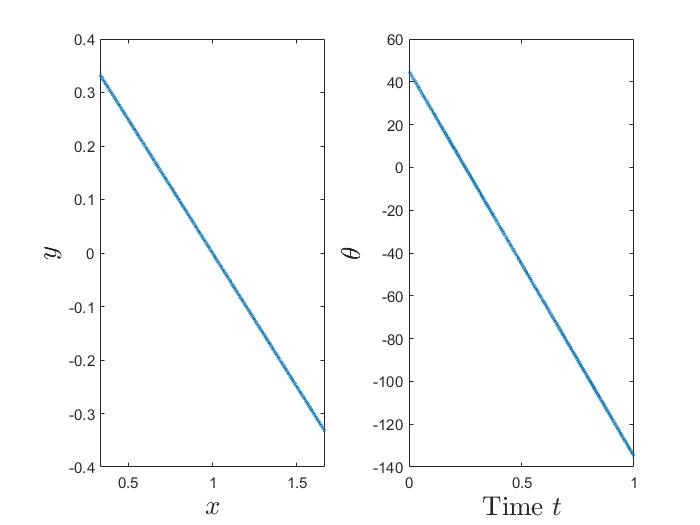}
    \caption{Trajectory of CoM; Angle of CoM (unit: degrees)}
    \label{fig:threeagent}
\end{figure}

Then we provide a three-agent system.
The terminal time is $t_f=1$.
Solving this problem, we got optimal cost $J=9.9131$.
Fig. \ref{fig:threeagent_traj} and Fig. \ref{fig:threeagent} show that the CoM moves with a constant speed and constant angular velocity.
\subsection{Example 3:  Four-agents model}
\begin{figure}[!htp]
    \centering
    \includegraphics[width=\linewidth]{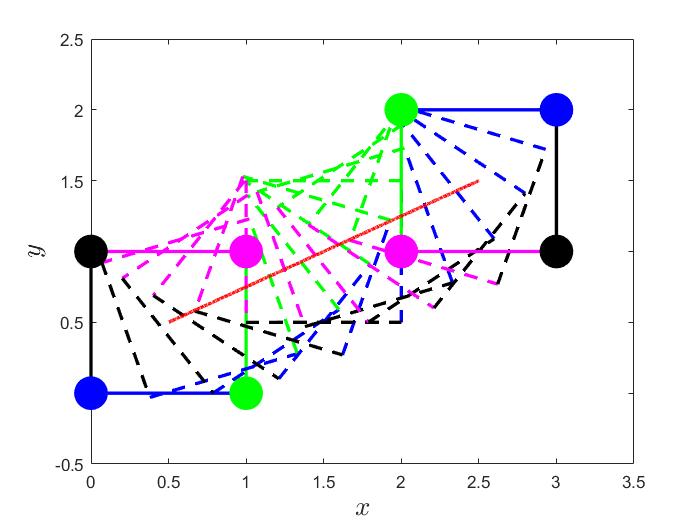}
    \caption{Trajectory: blue, black, green, magenta dotted line (dots): the trajectory of the system (agents); red line: the trajectory of the center of mass}
    \label{fig:fourAgentTraj}
\end{figure}

\begin{figure}[!htp]
    \centering
    \includegraphics[width=\linewidth]{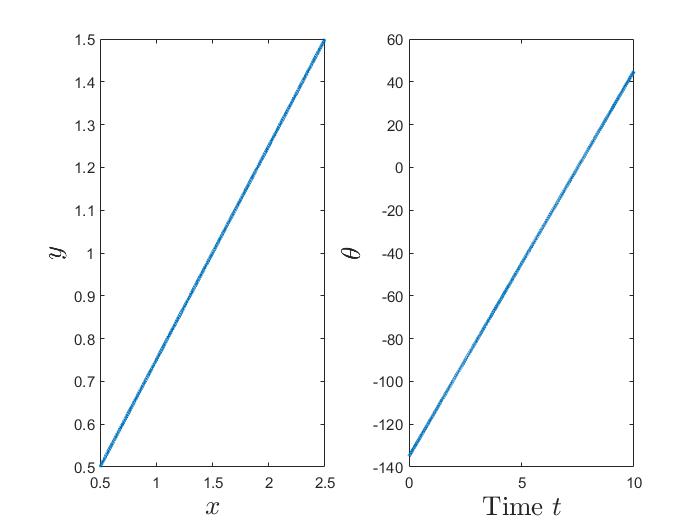}
    \caption{Trajectory of CoM; Angle $\theta$ between CoM and the blue agent}
    \label{fig:four_agent_CoM}
\end{figure}

Finally, we provide a simulation for a four-agents system.
For this system, we need five constraints to keep the rigidity of system.
We set $t_0=0$, $t_f=10$.
We found the optimal trajectory of this rigid body as Fig \ref{fig:fourAgentTraj} shows.
The total cost is $J =  1.9869$.
The center of mass is moving in a line which implies that the optimal control for each agent is the optimal control of the center of mass.
Fig. \ref{fig:four_agent_CoM} shows the trajectory of the CoM and the angle between CoM and the blue agent (unit: degrees).

\subsection{Application of this equivalence}
The above examples show that by using this equivalence between the multi-agent system and the center of mass, we actually simplified the original multiple-state optimal control problem into a three-state optimal control problem.
With the information of the initial angle and position from the center of mass to each agent known, we can then recover the optimal controller for each agent.
This application of equivalence significantly reduces the complexity of solving this optimal control problem.
Take the four-agent system as an example.
In this system, the original optimal control problem relates to eight states, eight controllers, and five equality constraints.
However, if we only consider the path planning for CoM, the dimension of the state space is 3 (i.e., $r_c \in \mathbb{R}^2$ and $\theta \in \mathbb{R}^1$) and 3 controllers (i.e., $u_c\in \mathbb{R}^2$ and $\omega \in \mathbb{R}^1$) without other constraints.
We can thus solve the optimal control problem for the center of mass, then recover the optimal controller for each agent.

\section{Conclusion}\label{sec:conclusion}
In this work, we studied the optimal path planning of homogeneous multi-agent systems with rigid formation.
We proved the equivalence of the original multi-agent system optimal control problem to the new single-CoM optimal control problem.
Finally, we provided some simulation results to illustrate our theoretical results.
In the future, we can extend our work to heterogeneous multi-agent systems which means agents may have different dynamics.
One other extension lies in higher dimensional ($\geq 3$) control.
One other possible extension is to add constraints that do not make the entire formation rigid.
Another interesting extension could be the same path planning problem with minimized energy in a heterogeneous terrain proposed in \cite{hetero}, but with multiple agents.

\bibliography{ifacconf}             
                                                   







\end{document}